\documentclass[12pt]{article}
\usepackage[utf8]{inputenc}
\usepackage[T1]{fontenc}
\usepackage{authblk}
\usepackage{amsmath}
\usepackage{amssymb}
\usepackage{amsbsy}
\usepackage{graphicx}
\usepackage{setspace}
\usepackage{tocloft}
\usepackage{varwidth}
\usepackage{algpseudocode}
\usepackage{listings}
\usepackage{hyperref}
\usepackage{newtxmath}
\usepackage{natbib}
\newtheorem{theorem}{Theorem}

 \usepackage{verbatim}
 \usepackage[all]{xy}
 \xyoption{arc}
 \usepackage{xkeyval} 
 \usepackage{tikz}
 \usepackage{etoolbox}
 \usepackage{multirow}
 \usepackage{caption}
\usetikzlibrary{intersections}

\date{}
 \title{sJIVE: Supervised Joint and Individual Variation Explained}
 \author[1]{Palzer EF}
 \author[2]{Wendt C}
 \author[3]{Bowler R}
 \author[4]{Hersh CP}
 \author[1]{Safo SE}
 \author[1]{Lock EF}

 \affil[1]{Division of Biostatistics, University of Minnesota, Minneapolis, 55455, USA}
 \affil[2]{Division of Pulmonary, Allergy and Critical Care, University of Minnesota, Minneapolis, 55455, USA}
 \affil[3]{Division of Pulmonary, Critical Care and Sleep Medicine, Department of Medicine, National Jewish Health, Denver, CO, USA}
 \affil[4]{Channing Division of Network Medicine, Brigham and Women's Hospital, Harvard Medical School, Boston, MA, USA}

\begin{document}
 \maketitle
 
\vspace{-1.2cm}
\abstract{
Analyzing multi-source data, which are multiple views of data on the same subjects, has become increasingly common in molecular biomedical research. Recent methods have sought to uncover underlying structure and relationships within and/or between the data sources, and other methods have sought to build a predictive model for an outcome using all sources. However, existing methods that do both are presently limited because they either (1) only consider data structure shared by all datasets while ignoring structures unique to each source, or (2) they extract underlying structures first without consideration to the outcome. We propose a method called supervised joint and individual variation explained (sJIVE) that can simultaneously (1) identify shared (joint) and source-specific (individual) underlying structure and (2) build a linear prediction model for an outcome using these structures. These two components are weighted to compromise between explaining variation in the multi-source data and in the outcome.  Simulations show sJIVE to outperform existing methods when large amounts of noise are present in the multi-source data.  An application to data from the COPDGene study reveals gene expression and proteomic patterns that are predictive of lung function. Functions to perform sJIVE are included in the {\tt r.jive} package, available online at \href{http://github.com/lockEF/r.jive}{http://github.com/lockEF/r.jive}.}

\section{Introduction}

Access to multiple sets of characteristics, or views, on the same group of individuals is becoming increasingly common. Each view is distinct but potentially related to one another. For example, Genetic Epidemiology of COPD (COPDGene) \citep{COPDGene} is an ongoing study that is obtaining numerous views on the same group of people including clinical, RNA sequencing (RNAseq), and proteomic data that aims to uncover the pathobiology in COPD. Such datasets, referred to as multi-view or multi-source data, have fueled an active area of statistical research to develop methods that [1] seek to uncover the relationships between and within each dataset, and/or [2] use multi-source data to create prediction models.

Underlying signals and associations that are shared across all datasets are called joint structures. Numerous methods can identify joint structure, many of which are extensions of canonical correlation analysis (CCA) \citep{CCA_norm, CCA_suff, CCA_FDR,rsCCA}. CCA constructs canonical variates, linear combinations of the variables, for each dataset such that the correlation between two datasets' variates is maximized \citep{CCA}. Extensions of CCA \citep{rodosthenous2020integrating} allow for more than two datasets, incorporate penalties for sparsity and regularization, or may maximize other measures of association such as covariance. Further extensions of CCA have incorporated supervision into the method in order for an outcome to influence the construction of the canonical variates \citep{rodosthenous2020integrating, CoRe, CVR, supCCA, SAR_sCCA}. By choosing canonical variates associated with an outcome, these CCA-based methods can build a prediction model based on the joint structure in the data.
Other methods have combined CCA with discriminant analysis to build similar joint prediction models.  In particular, joint association and classification analysis (JACA)\citep{JACA} and sparse integrative discriminant analysis (SIDA) \citep{SIDA} combine linear discriminant and canonical correlation analyses to identify latent vectors that explain the association in multi-source data and that optimally separate subjects into different groups. 

A limitation of predictive methods based on CCA is that they seek signal that is shared across all data sources, while relevant signal can also be specific to a single data source.  One way to capture all variation in a dataset, shared or not shared, is by principle components analysis (PCA) \citep{PCA}. PCA uses singular value decomposition (SVD) for dimension reduction while maximizing the variance. The low-rank output from PCA can then act as the design matrix in a regression framework for prediction \citep{bair2006prediction}. In the multi-source setting, PCA for prediction can be applied to each dataset individually, or to a concatenated matrix of all datasets. However, often multi-source datasets will contain both joint (i.e., shared) and individual (i.e., source-specific) signal, which PCA does not distinguish. Since 2010, several more flexible methods have been developed to capture both joint and individual structures within multi-source data \citep{DISCO, MOFA, GCA, GIPCA, SLIDE}. In particular, joint and individual variation explained (JIVE) \citep{JIVE, o2016r} is an extension of PCA or the singular value decomposition (SVD) that decomposes the data into low-rank and orthogonal joint and individual components. JIVE and the other methods referenced are solely exploratory, in that they do not inherently involve prediction or supervision for an outcome. 

Few methods follow a supervised approach using both joint and individual signals from the data. Supervised integrated factor analysis (SIFA) \citep{SIFA} uses the outcome to supervise the construction of the joint and individual components, but does not have a predictive model for the outcome. Recently, a bayesian method for joint association and prediction that incorporates prior functional  information was proposed \citep{chekouo2020bayesian}. JIVE-predict \citep{JIVE.pred} uses the joint and individual scores from the JIVE output to formulate a predictive model, and this approach has successfully been applied to COPD \citep{JIVE_COPD} and brain imaging data \citep{JIVE_image}. However, this 2-step approach always determines the joint and individual components without consideration for the outcome, which may hinder the predictive accuracy of the method.

In this paper, we propose supervised joint and individual variance explained (sJIVE) to find joint and individual components while simultaneously predicting a continuous outcome. This 1-step approach allows for joint and individual components to be influenced by their association with the outcome, explaining variation in both the multi-source data and the outcome in a single step.  

The rest of the article is organized as follows. In Section \ref{section:2}, we review JIVE and introduce sJIVE's methodology and estimation technique. In Section \ref{section:3}, we compare sJIVE to JIVE and JIVE-predict under a variety of conditions. In Section \ref{section:4}, we compare sJIVE to existing methods. In Section  \ref{section:5}, we apply sJIVE and other competing methods to COPD data, and conduct a pathway analysis to interpret the results results. In Section  \ref{section:6}, we discuss limitations and future research. Additional methodology details and simulation results can be found in the appendix. 

\section{Method and Estimation}
\label{section:2}

\newcommand{\y}{\mathbf{y}}
\newcommand{\Y}{\mathbf{y}}
\newcommand{\X}{\mathbf{X}}
\newcommand{\U}{\mathbf{U}}
\newcommand{\W}{\mathbf{W}}
\renewcommand{\S}{\mathbf{S}}
\newcommand{\Ta}{\boldsymbol{\theta}_1}
\newcommand{\Tb}{\boldsymbol{\theta}_{2i}}
\newcommand{\E}{\mathbf{E}}

\subsection{Framework and Notation}
Throughout this article, bold uppercase letters $(\mathbf{A})$ will denote matrices, bold lowercase letters $(\mathbf{a})$ will denote vectors, and unbolded lowercase letters $(a)$ will denote scalars. Define the squared Frobenius norm for an $m \times n$ matrix $\mathbf{A}$ by $\Vert \mathbf{A}
\Vert^2_F=\sum_{i=1}^m \sum_{j=1}^n a_{ij}^2$.  Define the row space of a matrix $\mathbf{A}$ by $\mbox{row}(\mathbf{A})$. 

For our context, there are $k$ views of data on the same $n$ subjects.  Each dataset $\X_i$, $i=1,\dots,k$ must contain complete data and are of size $p_i \times n$ where $p_i$ is the number of variables in the $i$th dataset. By default each row, or variable, in $\X_i$ $i=1,\dots,k$ is centered and scaled to have mean 0 and variance 1. Let $\X$ (without a subscript) represent the concatenation of all views of data, $\X=[\X_1^T \cdots \X_k^T]^T$.  Let $\Y$ be a length $n$ outcome vector, which by default we center and scale to have mean 0 and variance 1.  Our goal is to identify underlying structures within and between $\{\X_i\}_{i=1}^k$ and simultaneously build a predictive model for $\y$ from $\X$.  


\subsection{Review of JIVE and JIVE-predict}
\label{section:2.1}

We first review JIVE and JIVE-predict before explaining our proposed method. When $k=2$, JIVE decomposes $\X$ as follows: 
\begin{align*}
    \X_1 &= \U_1 \S_J + \W_1 \S_1 + \E_1 \\
    \X_2 &= \U_2 \S_J + \W_2 \S_2 + \E_2, 
\end{align*}
where $\U_i$ and $\S_J$ make up the joint component and $\W_i$ and $\S_i$ are the individual components with some error matrix, $\E_i$, $i=1,2$. In order to uniquely define the decomposition and distinguish the joint and individual signals, the joint and individual components in the $i$'th dataset are orthogonal to each other. These structures are represented by reduced-rank matrices with the ranks $r_J$, $r_1$, and $r_2$ where the ranks are pre-determined by a permutation \citep{JIVE} or Bayesian information criterion \citep{o2016r} approach, and $r_J$ is the rank of the joint structure while $r_i$ $i=1,\dots,k$ are the ranks of the individual structures. The loadings of the $i$th dataset,  $\U_{i\lbrace p_i \times r_J \rbrace}$ and $\W_{i\lbrace p_i \times r_i \rbrace}$, map the $p_i$ predictors to the low-rank subspace. Conversely, the score matrices, $\S_J$, $\S_1$, and $\S_2$, map the data from the low-rank subspace to the $n$ subjects. Note that $\S_J$, the joint scores, are the same for $\X_1$ and $\X_2$ and thus capture structure in the samples that is shared across the sources.

JIVE-predict uses the JIVE output for prediction of an outcome, $\Y$. The scores, $\S_J$ and $\S_i$ $\forall i=1,\dots, k$, can be used as a design matrix for a regression model. For example, when $k=2$, we can model $\Y$ linearly by
\begin{align*}
    \Y = \boldsymbol{\theta}_1 \S_J + \boldsymbol{\theta}_2 \S_1 + \boldsymbol{\theta}_3 \S_2 + \E_Y,
\end{align*}
where $\boldsymbol{\theta}_1$, $\boldsymbol{\theta}_2$, and $\boldsymbol{\theta}_3$ are vectors of length $r_J$, $r_1$, and $r_2$ respectively such that the total number of regression parameters is the sum of the ranks. Since JIVE calculates the joint and individual components without consideration to an outcome, the predictive accuracy of this two-stage approach may be limited. Thus, our proposed method, sJIVE, allows for us to simultaneously construct joint and individual components while building a linear regression model. This allows for the components to be influenced by their association with $\y$, which could increase predictive accuracy.

\subsection{Proposed model and objective}
\label{section:2.2}

Consider the simple case when $k=2$. Then, $\X$ and $\y$ can be decomposed by
\begin{align*}
        \X_1 &= \U_1 \S_J + \W_1 \S_1 + \E_1 \\
    \X_2 &= \U_2 \S_J + \W_2 \S_2 + \E_2 \\
    \y &= \boldsymbol{\theta}_1 \S_J + \boldsymbol{\theta}_{21} \S_1 + \boldsymbol{\theta}_{22} \S_2 + \E_Y.
\end{align*}
Similar to JIVE-predict, the scores from the decomposition of $\X_1$ and $\X_2$ are used to predict $\Y$. However, the calculation of $\S_J$ and $\S_i$ $\forall i=1, \dots , k$ are influenced by their association with the outcome. This can be accomplished by minimizing the loss of both $\X$ and $\Y$ through the following optimization problem:
    \begin{alignat}{3}
        \begin{split}
         \underset{ \S_J, \Ta, \Tb, \U_i, \W_i, \S_i, i=1,...,k}{\operatorname{argmin}} \hspace{.2cm} \sum_{i=1}^k \eta \Vert \X_i - \U_i \S_J - \W_i \S_i  \Vert^2_F  \\
 + (1-\eta)  \Vert \Y  - \Ta \S_J - \sum_{i=1}^k \Tb \S_i   \Vert^2_F 
        \end{split}
        \label{eqn1}
    \end{alignat}
where $\eta$ is a weight parameter between 0 and 1 to signify the relative importance of $\X$ and $\Y$ when calculating the loss. The joint structure across all $k$ datasets is represented by $\U_i$ $i=i,\dots,k$ and $\S_J$. Unique structure from the $i$th dataset is accounted for by $\W_i$ and $\S_i$. Low-rank approximations are used for both the joint and individual components such that $\U_i$ is $p_i \times r_J$ and $\S_J$ is $r_J \times n$ for the joint components, and $\W_i$ is $p_i \times r_i$ and $\S_i$ is $r_i \times n$ for the $i$th individual component. Section \ref{section:2.4} will discuss how ranks are selected. The second expression in (\ref{eqn1}) allows for the joint and individual scores to linearly predict the outcome, $\Y$. The coefficients, $\Ta$ and $\Tb$  $i=1,\cdots,k$, are vectors with lengths equal to the joint and individual ranks. 


\subsection{Identifiability}
\renewcommand{\j}{\mathbf{j}}
\renewcommand{\a}{\mathbf{a}}
\newcommand{\J}{\mathbf{J}}
\newcommand{\A}{\mathbf{A}}
\newcommand{\row}{\mbox{row}}

Consider the sJIVE approximation without error, 
\begin{align}
        \tilde{\X}_i &= \J_i + \A_i  \; \; \text{for} \; i=1,\hdots,k \; \; \;  \text{and} \; \; \;  
        \tilde{\y} = \j_y+\a_y,
        \label{sJIVE_noerror}
\end{align}
where $\J_i = \U_i \S_J$, $A_i=\W_i \S_i$, $\j_y= \boldsymbol{\theta}_1 \S_J$ and $\a_y= \sum_{i=1}^k \boldsymbol{\theta}_{2i} \S_i$.  Let $\tilde{\X}=[\tilde{\X}_1^T \cdots \tilde{\X}_k^T]^T$, $\J=[\J_1^T \cdots \J_k^T]^T$, and $\A=[\A_1^T \cdots \A_k^T]^T$.  Theorem~\ref{thm:ident} describes conditions for the identifiability of the terms in \eqref{sJIVE_noerror}.

\begin{theorem}
\label{thm:ident}
Consider $\{\tilde{\X}_1,\hdots, \tilde{\X}_k, \tilde{\y}\}$ where $\tilde{\y} \in \row(\tilde{\X})$. There exists a uniquely defined decomposition~\eqref{sJIVE_noerror} satisfying the following conditions:
\begin{enumerate}
\item $\row(\J_i) = \row(\J) \subset \row(\tilde{\X}_k)$ for $i=1,\hdots, k$,
\item $\row(\J) \perp \row(\A_i)$  for  $i=1\hdots,k$, 
\item $\cap_{i=1}^k \row(\A_i)=\{\mathbf{0}\}$,  
\item $\j_y \in \row(\J)$ and $\a_y \in \row(\A)$. 
\end{enumerate}
\end{theorem}
A proof of Theorem~\ref{thm:ident} can be found in Appendix~\ref{section:7.1}.  The first three conditions are equivalent to those for the JIVE decomposition (see Lemma 1 for angle-based JIVE \citep{feng2018angle}), which similarly enforces orthogonality between the joint and individual structures (condition 2.). Condition (4.) implies that predictions for the outcome can be uniquely decomposed into the contributions of joint and individual structure.  If the row spaces of each $\A_i$ are linearly independent (rank$(A)=\sum _{i=1}^k r_i)$, then the contribution of individual structure for each data source,  $\theta_{2i} \S_{i}$, are also uniquely determined.   We identify more granular terms by the SVD of joint or individual structures: $[\U_1^T \cdots \U_k^T \; \boldsymbol{\theta}_1^T]^T$ are given by the normalized left singular vectors of $[\J^T \; \j_y^T]^T$ and $[\W_i^T \, \boldsymbol{\theta}_{2i}^T]^T$ are the normalized left singular vectors of $[\A_i^T \; \boldsymbol{\theta}_{2i} \S_{i}]$ for  $i \in 1, \dots, k$.

\subsection{Estimation}
\label{section:2.3}

Our optimization function assumes the weighting parameter $\eta$ is fixed, but generally its value is not predetermined for a given application. In practice, we recommend running sJIVE with 5-fold cross validation (CV) for a range of possible $\eta$ values, choosing the one with the lowest mean squared error (MSE) for the test set. When $\eta=1$, our objective reduces to that of JIVE and will produce identical results. When $\eta$ shifts away from 1 and closer to 0, the results from sJIVE will start to deviate from JIVE in order to minimize the squared error of the $\Y$ expression. This allows the joint and individual components to be selected based on their association with the outcome.

By distributing $\eta$ throughout the expression, let $\Tilde{\mathbf{X}}_i = \eta^{1/2} \mathbf{X}_i$, $\Tilde{\mathbf{U}}_i = \eta^{1/2} \mathbf{U}_i$, $\Tilde{\mathbf{W}}_i = \eta^{1/2} \mathbf{W}_i$, $\Tilde{\mathbf{y}} = (1-\eta)^{1/2} \mathbf{y}$, $\Tilde{\boldsymbol{\theta}}_1 = (1-\eta)^{1/2} \boldsymbol{\theta}_1$ and $\Tilde{\boldsymbol{\theta}}_{2i} = (1-\eta)^{1/2} \boldsymbol{\theta}_{2i}$ $\forall i \in 1, \dots , k$. Furthermore, let $\Tilde{\W} \S$ denote $[ (\Tilde{\W}_1 \S_1)^T$ $\cdots$ $(\Tilde{\W}_k \S_k)^T]^T$. Then, we can minimize equation (\ref{eqn1}) as described in Algorithm 1. \\

\textbf{\hspace{-.5cm} Algorithm 1: Pseudocode for sJIVE estimation once $\eta, r_J$, and $r_i$ is known $\forall i\in 1, \dots, k$} \\
\framebox{

\parbox[t][9cm]{12cm}{

\flushleft
\begin{enumerate} \vspace{-.2cm}
     \item Initialize $\Tilde{\U}_i, \Tilde{\W}_i, \S_i$, $\Ta$, $\Tb$ $\forall i \in 1, \cdots, k$
     \item Loop until convergence
\begin{enumerate}
     \item  $\S_J = \begin{bmatrix} \Tilde{\U} \\ \Ta \end{bmatrix}^T \Big(\begin{bmatrix} \Tilde{\X} \\ \Tilde{\Y} \end{bmatrix} - \begin{bmatrix}
\Tilde{\mathbf{W}} \S \\
 \sum_{i=1}^k \Tb \S_i
\end{bmatrix} \Big)$
\item $[\Tilde{\U}^T$ $\Ta^T]^T$ = the first $r_J$ left singular vectors from \\ SVD$ \Big( \begin{bmatrix} \Tilde{\X} \\ \Tilde{\Y} \end{bmatrix}^T
 - \begin{bmatrix} \Tilde{\W} \Tilde{\S} \\
 \sum_{i=1}^k \Tb \S_i \end{bmatrix} \Big)$
        \item for each dataset $i = 1,...,k$
        \begin{enumerate}
            \item $\S_i = \begin{bmatrix} \Tilde{\W}_i \\ \Tb \S_i  \end{bmatrix}^T \Big( \begin{bmatrix}  \Tilde{\X} \\ \Tilde{\Y} - \sum_{j \ne i} \Tilde{\boldsymbol{\theta}}_{2j} \S_j \end{bmatrix} -   \begin{bmatrix} \Tilde{\U}_i \\ 
            \Ta \end{bmatrix} \S_J \Big) P_J^C$
            \item $[\Tilde{\W}_i^T$ $\Tb^T]^T =$  the first $r_i$ left singular vectors from \\ SVD$\Big(  \begin{bmatrix}  \Tilde{\X} \\ \Tilde{\Y} - \sum_{j \ne i} \Tilde{\boldsymbol{\theta}}_{2j} \S_j \end{bmatrix} -   \begin{bmatrix} \Tilde{\U}_i \\ 
            \Ta \end{bmatrix} \S_J \Big) P_J^C$
        \end{enumerate}
        \item Check for convergence of $\big\Vert \begin{bmatrix} \Tilde{\X} \\ \Tilde{\Y} \end{bmatrix} - \begin{bmatrix} \Tilde{\U} \\ \Ta \end{bmatrix} \mathbf{S}_J - \begin{bmatrix}
\Tilde{\mathbf{W}} \S \\
 \sum_{i=1}^k \Tb \S_i
\end{bmatrix}
 \big\Vert^2_F $
\end{enumerate}
\end{enumerate}
} }

\vspace{.5cm}
 
After initializing the parameters, our model iteritively solves for the joint and individual components until convergence of our optimization function. In step 2c of Algorithm 1, the individual components are projected onto the orthogonal complement of the joint subspace, $P_J^C$, to retain orthogonality. After the algorithm converges, additional scaling is needed for the results to be identifiable. The joint loadings and regression coefficients are scaled such that $[\U_1^T \cdots \U_k^T$ $\boldsymbol{\theta}_1^T]^T$ has a squared Frobenius norm of 1. The joint scores, $\S_J$ absorb this scaling as to not change the overall joint effect, $\mathbf{J}$, where $\mathbf{J} = [\U_1^T \cdots \U_k^T$ $\boldsymbol{\theta}_1^T]^T \S_J$. The same scaling can be done for each individual effect such that $[\W_i^T$ $\boldsymbol{\theta}_{2i}^T]^T$ $\forall i \in 1, \cdots, k$ have a squared Frobenius norm of 1 and $\S_i$ absorbs this scaling for the $i$th dataset $i=1,\dots,k$. 

\subsection{sJIVE-prediction for a Test Set}

After running sJIVE, we may want to predict new outcomes with external data or with a test set. Let $m$ be the number of out-of-sample observations, and let us have complete data in $\textbf{X}^\ast_i$, $i=1,\dots,k$ for each of the $m$ samples. By extracting $\hat{\textbf{U}}_i$, $\hat{\textbf{W}}_i$, $\hat{\boldsymbol{\theta}}_1$, and $\hat{\boldsymbol{\theta}}_{2i}$ $\forall i=1,\dots,k$ from the fitted sJIVE model and treating them as fixed, we solve the following optimization function to obtain joint and individual scores for the new data,
\begin{align}
\label{eqn2}
\underset{ \S_J, \S_i, i=1,...,k}{\operatorname{argmin}} \hspace{.3cm} &\sum_{i=1}^k \Vert \X_i^\ast - \hat{\U}_i \S_J - \hat{\W}_i \S_i  \Vert^2_F 
\end{align}
where $\S_J$ is $r_J \times m$, and $\S_i$ is $r_i \times m$, $i=1,\dots,k$. We can iteratively solve for $\S_J$ and $\S_i$, $i=1,\dots,k$ using the closed-form solutions,
\begin{align*}
    \hat{\S}_J &= \begin{bmatrix} \hat{\U}_1 \\
    \vdots \\ \hat{\U}_k \end{bmatrix}^T \Big(\begin{bmatrix} \X^\ast_1 \\
    \vdots \\ \X^\ast_k \end{bmatrix} - \begin{bmatrix}
\hat{\mathbf{W}}_1 \hat{\S}_1 \\
\vdots \\
\hat{\mathbf{W}}_k \hat{\S}_k \end{bmatrix} \Big) \\
\hat{\S}_i &= \hat{\W}_i^T (\X_i^\ast - \hat{\U}_i \hat{\S}_J) 
\end{align*}
Using the the newly-obtained scores, the fitted outcomes can be estimated by $\hat{\Y}^\ast = \hat{\boldsymbol{\theta}}_1 \hat{\S}_J + \sum_{i=1}^k \hat{\boldsymbol{\theta}}_{2i} \hat{\S}_i$ using $\hat{\boldsymbol{\theta}}_1$ and $\hat{\boldsymbol{\theta}}_{2i}$ $\forall i=1,\dots,k$ from the original sJIVE model.

\subsection{Rank Selection}
\label{section:2.4}

Choosing an appropriate reduced rank for the joint and individual components is necessary for optimal model performance.  By default, sJIVE selects ranks via a forward selection 5-fold cross validation (CV) approach. The ranks are iteratively added in order to minimize the average test MSE for $\y$. Once adding an additional rank fails to lower the MSE, the function stops and the ranks are recorded. For more details, see Appendix~\ref{section:7.2}. 

In contrast to this approach, JIVE uses a permutation approach to select ranks. For both approaches, the joint rank must be $\le min(n, p_1, \dots, p_k)$ and the individual rank for dataset $i$ must be $\le min(n, p_i)$ $\forall i=1,\cdots,k$. In section \ref{section:3.4}, we will compare the accuracy of sJIVE's CV approach to JIVE's permutation approach.

\section{Compare sJIVE to JIVE-predict}
\label{section:3}

\subsection{Simulation Set-up}
\label{section:3.1}

Since our work is an extension of JIVE and JIVE-predict, we will first assess how these models compare to supervised JIVE. Datasets were simulated to reflect the sJIVE framework such that the true joint and individual components are known. 

Data were simulated as follows: Predictors $\X_1$ and $\X_2$, with $p_1=p_2=200$ rows and $n=200$ columns, were generated such that the true reduced joint and individual ranks are all equal to $1$. Both the joint and individual components contribute equally to the total amount of variation in $\X$ with varying levels of noise. Similarly, both the joint and individual scores contain equal signal to the composition of $\Y$. We assessed the performance of each model by the MSE of an independent test set of $n=200$. Each scenario was simulated 10 times. For additional details about the data generation, see Appendix~\ref{section:7.3}.  

 For Sections~\ref{section:3.2} and~\ref{section:3.3} we use the true ranks, and we compare the accuracy of rank selection approaches for sJIVE and JIVE in Section~\ref{section:3.4}. We varied the amount of noise in $\X$ and in $\Y$ across simulations. For example, when $\X$ error is 10\%, 10\% of the variation in $\X$ is due to error, and the joint and individual components can account for 90\% of the total variation in $\X$. Each test MSE is the average of the 10 simulations, and the final column records how often sJIVE outperformed JIVE-predict. 
 
 \subsection{Comparing Test MSE}
 \label{section:3.2}
 
 The results displayed in Table \ref{tab:JIVE_MSE} show that when there is a relatively small amount of noise in $\X$ or a relatively large noise in $\Y$, sJIVE tends to perform similar to JIVE. However, when the amount of noise in $\X$ is between 30\% and 99\% with low error in $\Y$, sJIVE consistently outperforms JIVE-predict. In these cases, the large amount of noise in $\X$ makes it difficult for JIVE to capture the signal. sJIVE incorporates $\Y$ when determining the joint and individual components, which helps uncover these signals when there is error in $\X$.  When the amount of error in $\X$ is greater than 99\%, neither sJIVE nor JIVE-predict are able to detect the signal in $\X$.

 \subsection{Recovering True Structures}
\label{section:3.3}

Similarly, we tested how well sJIVE and JIVE were able to reconstruct the true joint and individual components. Accuracy of each component was summarized by the standardized squared Frobenius norm difference, e.g.,  $||(\hat{\mathbf{J}} - \mathbf{J})||_F^2 / ||\mathbf{J}||_F^2 $ where $\hat{\mathbf{J}}$ is the estimated joint component and $\mathbf{J}$ is the true joint component. A similar measure can be obtained for the accuracy of each individual component. Both sJIVE and JIVE struggled with finding the true components when error in $X$ was over 99\% (Table \ref{tab:JIVE_norm}). When comparing the two methods to each other, sJIVE tended to identify the individual components more accurately than JIVE for all levels of error in $\X$ and $\Y$. However, when the error in $\X$ was greater than 99\%, sJIVE tended to more accurately identify the individual components, while JIVE was able to more accurately identify the joint component.

In terms of both test MSE and identifying the true components, sJIVE and JIVE-predict failed to perform well in cases where $\X$ error was greater than 99\%. The largest eigenvalue of the signal in $\X$ drops below that of the noise when the error in $\X$ is above 97\%. When this occurs, statistical models tend to struggle at identifying the signal; rather, they capture a mix of the true signal and the noise. However, when $\X$ error was 99\%, sJIVE continued to perform mildly better in terms of test MSE compared to JIVE-predict, especially when error in $\Y$ was low, suggesting that sJIVE can more effectively separate noise from signal in $\X$ when large amounts of noise are present.

 \subsection{Comparing Rank Selection}
\label{section:3.4}

In addition to the simulations with known rank, we also compared the rank selection techniques of sJIVE and JIVE. JIVE uses a permutation approach while sJIVE uses a forward-selection CV method. For sJIVE's CV approach, all ranks are initially set to zero. Ranks are added if the additional rank results in the largest reduction in test MSE after running 5-fold CV. We simulated all combinations of joint and individual ranks of size 1 to 4 with defaults $n=100$, $k=2$, $p_1=p_2=100$ for each dataset, joint and individual components have equal weight, 50\% error in $\X$, and 10\% error in $\y$. Table \ref{tab:JIVE_rank} shows the percent of the time when sJIVE and JIVE were able to accurately specify the ranks. When the true joint rank was equal to 1, JIVE's permutation approach tended to perform equal or better than sJIVE's CV approach. However, when the true joint rank was greater than 1, JIVE rarely identified the true joint rank while sJIVE selected the true joint rank about 20\% of the time. In total, sJIVE  correctly identified all 3 ranks 2\% of the time, while JIVE had a 10\% chance. 

In Table \ref{tab:JIVE_rank2}, we further look at the probability of over- and under-estimating the ranks. JIVE's permutation method regularly underestimated the joint rank (80.5\%) and overestimated the individual ranks (58.5\% and 50.2\% for datasets $\X_1$ and $\X_2$ respectively). Additionally, JIVE correctly specified the joint rank about 20\% of the time across all simulations and 40\% of the time for each individual rank. In contrast, the sJIVE's probability of overestimating, underestimating, or correctly specifying the joint rank was similar to that of the individual ranks. sJIVE was twice as likely to underestimate each of the ranks compared to overestimate, and had about a 25\% chance of specifying any rank correctly.

\section{Compare sJIVE to Other Methods}
\label{section:4}

Next, we compared sJIVE to not only JIVE-predict, but also to concatenated PCA (i.e., PCA of the concatenated data), individual PCA (i.e., PCA of each dataset individually), and CVR \citep{CVR}. The data was generated in the same manner as earlier simulations. Our default parameters included $k=2$ datasets, all ranks set to 1, and equal contribution of the joint and individual components. Additionally, we set $\X$ error to 90\%, $\Y$ error to 1\%, $n=200$, and $p_1=p_2=200$. We tested the following 8 scenarios: [1] the listed default parameters, [2] increasing the number of predictors in each dataset, $p$, to 500, [3] increasing the error in $\X$ to 99\%, [4] increasing the number of datasets, $k$, to 4, [5] increasing the joint signal to be 20 times that of the individual signals, [6] increasing the individual signals to be 20 times greater than the joint signal, [7] increasing all ranks to be 10, and lastly, [8] increasing all ranks to be 10, but only allowing the first rank to be predictive of $\Y$. Though CVR can be extended to $k \ge 2$ datasets, its R package \citep{CVRpackage} only works for exactly 2 datasets. Thus, we could not obtain a test MSE for CVR when $k$ was greater than 2. Each scenario was run 10 times and the percent of time when sJIVE performed the best was recorded. 

The results can be found in Table \ref{tab:Compare}. sJIVE outperformed all other methods at least 80\% of the time for 6 of the 8 scenarios, with the exceptions of when a large joint rank or large individual rank are present. For a large joint rank, concatenated PCA marginally outperformed the other methods. Concatenated PCA only looks at the joint signal, ignoring the individual signals, which allows for better predictive accuracy in the presence of large joint effects. When there was a large individual signal, sJIVE and JIVE-predict performed equally well. Increasing the error in $\X$ to 99\% resulted in the largest improvement in sJIVE compared each of the other methods, but sJIVE only saw a modest increase in predictive accuracy in the first two scenarios. 

\section{Application to COPDGene Data}
\label{section:5}

Genetic Epidemiology of COPD (COPDGene) is a multi-center longitudinal observational study to identify genetic factors associated with chronic obstructive pulmonary disease (COPD). To attain this goal, multiple views of data were collected on the same group of individuals including RNA sequencing (RNAseq) data \citep{RNAseq} and proteomic data \citep{prot1, prot2}. Additionally, spriometry was performed to calculate a percent predicted forced expiratory volume in 1 second (FEV1\% predicted, or FEV1pp) value, a measure of pulmonary function which is significantly lower in COPD patients compared to their healthy counterparts. 

Our goal is to uncover underlying shared and individual structures from the RNAseq and proteomic datasets while simultaneously using these structures to predict pulmonary function. We have complete data on 359 participants for 21,669 RNAseq targets, 1,318 proteins, and a value for FEV1pp. We compare sJIVE to JIVE-predict, concatenated PCA, individual PCA, and CVR. The data were split into 2/3 training set and 1/3 test set with accuracy measured by test MSE, and ranks were determined by JIVE's permutation approach.

The results for each model can be found in Table \ref{tab:COPD}. Of the 6 methods tested, sJIVE had the lowest test MSE at 0.6980.  This implies that over 30\% of the variation in FEV1pp can be explained by the gene expression and proteomic profile. This result is substantially better than each of the PCA methods and CVR, though JIVE-predict had a similar test MSE of 0.6991. In terms of computation time on a 2.4 GHz computer with 8 GB RAM, sJIVE took slightly over an hour to run, which is almost twice as long as JIVE-predict, but over 100 times quicker than CVR. In high-dimensional settings, JIVE and sJIVE map their data to smaller dimensions to increase computational efficiency. This allows for both methods to handle larger datasets. See Appendix~\ref{section:6.4} for more details on this \emph{a priori} data compression. Overall, sJIVE results in only a modest increase in accuracy compared to JIVE-predict for the COPDGene data, but a substantial increase in accuracy compared to CVR and PCA approaches.

\begin{figure}
    \centering
    \includegraphics[width=12cm]{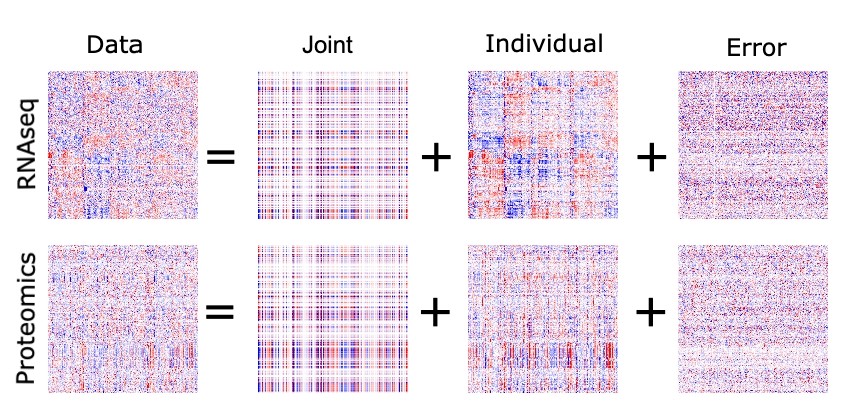}
    \caption{Heatmap of the joint and individual structure within the COPDGene dataset.}
    \label{fig:COPD_heatmap}
\end{figure}

\begin{figure}
    \centering
    \includegraphics[width=9cm]{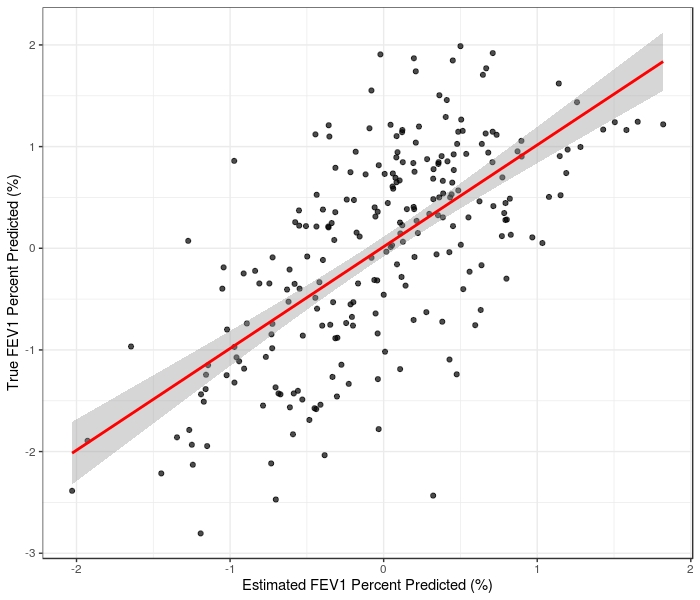}
    \caption{Comparing true FEV1 percent predicted values to estimated values from sJIVE output.}
    \label{fig:COPD_pred}
\end{figure}

In addition to testing the predictive accuracy of sJIVE, we further investigated the fit of the estimated model The underlying data structures are graphically displayed as heatmaps in Figure \ref{fig:COPD_heatmap}, with strong positive and negative values in red and blue, respectively. To further assess the predictive model, Figure \ref{fig:COPD_pred} compares the true FEV1pp values to the estimated ones, showing that a linear fit is reasonable and that there is no systematic over- or underestimation of FEV1pp.  

The results of the predicted model are further summarized in Table \ref{tab:COPD_sJIVE} to assess the effect of the joint and individual components on FEV1\% predict, and we use an F-test to assess the significance of each component in the multivariate model. JIVE's permutation method selected a joint rank of 1, and individual ranks of 27 and 24 for the RNAseq and proteomic data respectively. The single joint rank accounts for 6.9\% of the total variation in FEV1\% predicted $(p=0.001)$. As for the individual effects, proteomics account for 33.5\% of the variation $(p<0.001)$ and RNAseq accounts for 19.1\%. However, the individual effect of RNAseq failed to reach statistical significance $(p=0.217)$.  Our proteomic data exhibited a large, significant association with FEV1pp after removing the joint structure while RNAseq failed to attain significance beyond its contribution to the joint component. This suggests that post-transcriptional regulation or other factors unique to the proteome may influence lung function.  

\begin{figure}
    \centering
    \includegraphics[width=12cm]{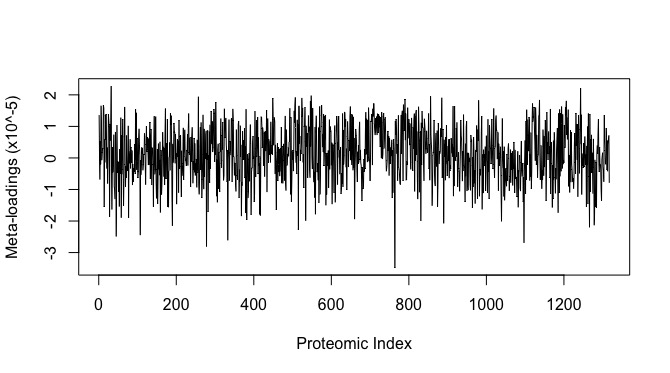}
    \caption{Meta-loadings from the sJIVE result for the proteomic dataset.}
    \label{fig:COPD_meta}
\end{figure}

To further investigate the proteomic results, we conducted a pathway analysis using the WEB-based GEne SeT AnaLysis Toolkit (WebGestalt) \citep{WebGestalt}. Using the results from sJIVE, we calculated the meta-loadings for each predictor in a similar manner to that in \citet{JIVE_COPD}, by taking the sum across the joint and individual loadings, $\U_i$ and $\W_i$, weighted by their regression coefficients. The meta-loadings for the proteomic dataset can be found in Figure \ref{fig:COPD_meta}. The top 20\% of absolute meta-loadings were used to perform an over-representation enrichment analysis with KEGG pathway database and the top 10 pathways are shown in Table \ref{tab:COPD_pathway}.  The following were found to be statistically significant pathways (all p<0.001): Glucagon signaling pathway, dopaminergic synapse, cholinergic synapse, wnt signaling pathway, B cell receptor signaling pathway, chemokine signaling pathway, AGE-RAGE signaling pathways in diabetic complications, VEGF signaling pathway, circadian entrainment, and insulin signaling pathway. The false discovery rate (FDR) for these 10 pathways remained under 0.005. Three of these pathways have been mechanistically linked to COPD, specifically the emphysema phenotype. RAGE or receptor for advanced glycosylation end product receptor has been identified as both a biomarker and mediator of emphysema \citep{COPD_ref1, COPD_ref2, COPD_ref3}. Similarly, VEGF has been mechanistically linked to emphysema and sputum VEGF levels are reciprocally related to the level of COPD \citep{COPD_ref4, COPD_ref5}. The Wnt signaling pathway is associated with aging and down-regulation of this pathway in human airway epithelium in smokers is associate with smoking and COPD \citep{COPD_ref6, COPD_ref7}. Importantly, this approach has the potential to identify and link novel pathways associated with COPD, opening new avenues for research. 

\section{Summary and Discussion}
\label{section:6}

We have proposed sJIVE as a one-step approach to identify joint and individual components in multi-source data that relate to an outcome, and use those components to create a prediction model for the outcome. This approach facilitates interpretation by identifying concordant and complementary effects among the different sources, and has competitive predictive performance.  When comparing sJIVE to a similar two-step approach, JIVE-predict, sJIVE performed best in the presence of large amounts of error in the multi-source data $\X$ when the error in $\Y$ was relatively small. Even in scenarios when the largest eigenvalue of the signal is slightly smaller than that of the noise, sJIVE was able to capture the signal better than JIVE-predict.  When comparing sJIVE to principal components and canonical correlation based approaches, sJIVE tended to perform best in almost all scenarios tested. 

When applying our method to the COPDGene data, sJIVE and JIVE-predict also outperformed the other methods. Though sJIVE resulted in the lowest test MSE, the gain in accuracy between the two methods was marginal in this case. sJIVE found significant associations between the proteomic data and FEV1\% predicted, as well as in the joint effect of proteomic and RNAseq data. After conducting a pathway analysis of the proteomic results, we uncovered 3 pathways that had previously been mechanistically linked to COPD, as well as additional pathways that could benefit by future research.

 Our method has some limitations and avenues for future work. While our simulations demonstrate its good performance when the correct ranks are selected, the correct ranks may not be selected in practice. Depending on the application, JIVE's permutation method or sJIVE's CV method for rank selection may be preferred, but neither are ideal. Future research can explore and compare alternative approaches to determine the ranks. Other methods of scaling, such as scaling each $\X_i$ to have a Frobenius norm of 1, have been used, which is different from our method that scales each predictor to have variance 1. Moreover, sJIVE treats all predictors and the outcome as continuous. A useful extension to our method would be to allow for a binary outcome, or other distributional forms. Additionally, our method does not allow for missing data, so missing values must be imputed or the entire observation must be removed. Lastly, sJIVE does not explicitly capture signal that is shared between some, but not all, data sources. Extensions that allow for partially-shared structure, such as in the SLIDE method \citep{SLIDE}, would allow for more flexibility. 

\section*{Acknowledgements}

The views expressed in this article are those of the authors and do not reflect the views of the United States Government, the Department of Veterans Affairs, the funders, the sponsors, or any of the authors’ affiliated academic institutions.

\section*{Funding}

This work was partially supported by grants R01-GM130622 and 5KL2TR002492-03 from the National Institutes of Health and by Award Number U01 HL089897 and Award Number U01 HL089856 from the National Heart, Lung, and Blood Institute. The content is solely the responsibility of the authors and does not necessarily represent the official views of the National Heart, Lung, and Blood Institute or the
National Institutes of Health.

COPDGene is also supported by the COPD Foundation through contributions made to an Industry Advisory Board comprised of AstraZeneca, Boehringer-Ingelheim, Genentech, GlaxoSmithKline, Novartis, Pfizer, Siemens, and Sunovion.

\bibliographystyle{natbib}
\bibliography{ref.bib}   

\newpage
\onecolumn

\section*{Figures and Tables}
\label{section:7}

\begin{table}[h!]
\centering
\footnotesize
\begin{tabular}{ccccccc}
  \hline
 X & Y& sJIVE& JIVE-predict &  \% of Time& Eigenvalue & Eigenvalue \\
Error & Error &  MSE &  MSE & sJIVE wins & of X signal & of X error \\
  \hline
0.10 & 0.10 & 0.1068 & 0.1063 &  40 & 154.46 & 8.85 \\ 
  0.10 & 0.30 & 0.3269 & 0.3280 & 70 & 153.52 & 8.83 \\ 
  0.10 & 0.50 & 0.5160 & 0.5147 &  60 & 154.04 & 8.81 \\ 
  0.10 & 0.70 & 0.7138 & 0.7160 & 90 & 154.22 & 8.88 \\ 
  0.10 & 0.90 & 0.9002 & 0.8990 & 50 & 155.70 & 8.83 \\ 
  0.30 & 0.10 & 0.1385 & 0.1394 &  70 & 137.05 & 15.34 \\ 
  0.30 & 0.30 & 0.3143 & 0.3145 &  60 & 136.52 & 15.38 \\ 
  0.30 & 0.50 & 0.5620 & 0.5619 &  50 & 136.90 & 15.25 \\ 
  0.30 & 0.70 & 0.6901 & 0.6888 &  50 & 135.98 & 15.34 \\ 
  0.30 & 0.90 & 0.9620 & 0.9593 &  50 & 136.84 & 15.26 \\ 
  0.50 & 0.10 & 0.1439 & 0.1483 &  100 & 115.45 & 19.80 \\ 
  0.50 & 0.30 & 0.3393 & 0.3395 & 50 & 114.67 & 19.74 \\ 
  0.50 & 0.50 & 0.6211 & 0.5262 & 60 & 115.01 & 19.70 \\ 
  0.50 & 0.70 & 0.7334 & 0.7306 & 50 & 115.71 & 19.71 \\ 
  0.50 & 0.90 & 0.9559 & 0.9296 & 60 & 114.43 & 19.70 \\ 
  0.70 & 0.10 & 0.1900 & 0.2016 & 100 & 89.55 & 23.38 \\ 
  0.70 & 0.30 & 0.4142 & 0.3833 & 50 & 88.94 & 23.31 \\ 
  0.70 & 0.50 & 0.5771 & 0.5728 & 50 & 89.98 & 23.32 \\ 
  0.70 & 0.70 & 0.8704 & 0.7631 & 40 & 88.14 & 23.41 \\ 
  0.70 & 0.90 & 0.9268 & 0.9250 & 70 & 90.33 & 23.50 \\ 
  0.90 & 0.10 & 0.2176 & 0.2160 & 70 & 51.89 & 26.37 \\ 
  0.90 & 0.30 & 0.4256 & 0.4180 & 60 & 51.73 & 26.35 \\ 
  0.90 & 0.50 & 0.5949 & 0.5801 & 50 & 51.10 & 26.44 \\ 
  0.90 & 0.70 & 0.7556 & 0.7413 & 30 & 51.84 & 26.31 \\ 
  0.90 & 0.90 & 0.8940 & 0.8945 & 40 & 51.49 & 26.31 \\ 
  0.95 & 0.10 & 0.3237 & 0.3383 & 100 & 36.38 & 27.09 \\ 
  0.95 & 0.30 & 0.4261 & 0.4278 & 60 & 36.57 & 26.91 \\ 
  0.95 & 0.50 & 0.6681 & 0.6490 & 40 & 36.47 & 27.16 \\ 
  0.95 & 0.70 & 0.7945 & 0.7960 & 80 & 36.68 & 27.18 \\ 
  0.95 & 0.90 & 0.9240 & 0.9158 & 10 & 36.69 & 27.21 \\ 
  0.99 & 0.10 & 0.7776 & 0.8549 & 90 & 16.28 & 27.54 \\ 
  0.99 & 0.30 & 0.8672 & 0.9252 & 70 & 16.41 & 27.53 \\ 
  0.99 & 0.50 & 0.9613 & 0.9452 & 60 & 16.28 & 27.66 \\ 
  0.99 & 0.70 & 0.9566 & 0.9504 & 70 & 16.24 & 27.75 \\ 
  0.99 & 0.90 & 1.0027 & 0.9745 & 30 & 16.48 & 27.77 \\ 
  0.999 & 0.10 & 1.0278 & 1.0017 & 70 & 5.18 & 27.85 \\ 
  0.999 & 0.30 & 0.9990 & 0.9970 & 50 & 5.19 & 27.82 \\ 
  0.999 & 0.50 & 1.0401 & 0.9976 & 30 & 5.14 & 27.90 \\ 
  0.999 & 0.70 & 1.0128 & 1.0005 & 60 & 5.12 & 27.74 \\ 
  0.999 & 0.90 & 1.0023 & 0.9983 & 40 & 5.16 & 27.84 \\ 
   \hline
\end{tabular}
\caption{Comparing Test MSE of sJIVE, JIVE-predict, and concatenated PCA. Defaults: n=200 for training set, $n=200$ for test set, $k=2$, $p=200$ for each dataset, joint and individual components have equal weight, and all ranks are 1.}
\label{tab:JIVE_MSE}
\end{table}

\begin{table}
\centering
\footnotesize
\begin{tabular}{cc|ccc|ccc|ccc}
  \hline
  X  & Y &  & sJIVE & &  & JIVE & &  \% sJIVE & beats & JIVE \\ 
  Error & Error & J & A1 & A2 & J & A1 & A2 & J & A1 & A2 \\
    \hline
0.10 & 0.10 & 0.036 & 0.035 & 0.028 & 0.034 & 0.034 & 0.027 & 10 & 20 & 30 \\ 
  0.10 & 0.30 & 0.042 & 0.033 & 0.029 & 0.042 & 0.033 & 0.030 & 80 & 50 & 70 \\ 
  0.10 & 0.50 & 0.035 & 0.030 & 0.034 & 0.034 & 0.030 & 0.035 & 40 & 50 & 40 \\ 
  0.10 & 0.70 & 0.036 & 0.027 & 0.024 & 0.036 & 0.026 & 0.024 & 50 & 50 & 70 \\ 
  0.10 & 0.90 & 0.049 & 0.029 & 0.031 & 0.048 & 0.029 & 0.030 & 20 & 50 & 20 \\ 
  0.30 & 0.10 & 0.115 & 0.070 & 0.060 & 0.132 & 0.080 & 0.068 & 100 & 100 & 100 \\ 
  0.30 & 0.30 & 0.142 & 0.091 & 0.086 & 0.145 & 0.094 & 0.085 & 70 & 80 & 70 \\ 
  0.30 & 0.50 & 0.128 & 0.081 & 0.076 & 0.129 & 0.081 & 0.077 & 60 & 60 & 60 \\ 
  0.30 & 0.70 & 0.117 & 0.083 & 0.075 & 0.119 & 0.084 & 0.076 & 90 & 80 & 100 \\ 
  0.30 & 0.90 & 0.145 & 0.083 & 0.081 & 0.148 & 0.084 & 0.083 & 70 & 70 & 70 \\ 
  0.50 & 0.10 & 0.261 & 0.123 & 0.136 & 0.315 & 0.157 & 0.168 & 100 & 100 & 100 \\ 
  0.50 & 0.30 & 0.280 & 0.155 & 0.163 & 0.303 & 0.169 & 0.176 & 100 & 90 & 100 \\ 
  0.50 & 0.50 & 0.547 & 0.259 & 0.310 & 0.408 & 0.225 & 0.214 & 80 & 80 & 80 \\ 
  0.50 & 0.70 & 0.285 & 0.146 & 0.156 & 0.289 & 0.148 & 0.158 & 90 & 80 & 90 \\ 
  0.50 & 0.90 & 0.458 & 0.236 & 0.281 & 0.308 & 0.163 & 0.188 & 60 & 60 & 70 \\ 
  0.70 & 0.10 & 0.924 & 0.502 & 0.442 & 1.041 & 0.579 & 0.513 & 90 & 100 & 100 \\ 
  0.70 & 0.30 & 1.020 & 0.527 & 0.500 & 1.002 & 0.548 & 0.462 & 90 & 80 & 90 \\ 
  0.70 & 0.50 & 0.788 & 0.437 & 0.379 & 0.828 & 0.459 & 0.401 & 90 & 80 & 90 \\ 
  0.70 & 0.70 & 0.945 & 0.434 & 0.436 & 0.935 & 0.426 & 0.419 & 70 & 70 & 60 \\ 
  0.70 & 0.90 & 0.887 & 0.483 & 0.437 & 0.894 & 0.489 & 0.439 & 90 & 90 & 80 \\ 
  0.90 & 0.10 & 1.751 & 0.756 & 0.734 & 1.745 & 0.786 & 0.829 & 60 & 90 & 100 \\ 
  0.90 & 0.30 & 1.573 & 0.851 & 0.873 & 1.578 & 0.886 & 0.913 & 70 & 100 & 90 \\ 
  0.90 & 0.50 & 1.737 & 0.819 & 0.787 & 1.736 & 0.832 & 0.797 & 50 & 80 & 50 \\ 
  0.90 & 0.70 & 1.665 & 0.765 & 0.800 & 1.710 & 0.798 & 0.815 & 70 & 90 & 80 \\ 
  0.90 & 0.90 & 1.774 & 0.760 & 0.801 & 1.788 & 0.793 & 0.793 & 60 & 70 & 70 \\ 
  0.95 & 0.10 & 2.454 & 1.204 & 1.095 & 2.431 & 1.327 & 1.282 & 50 & 90 & 100 \\ 
  0.95 & 0.30 & 2.390 & 1.274 & 1.218 & 2.461 & 1.268 & 1.310 & 90 & 80 & 90 \\ 
  0.95 & 0.50 & 2.428 & 1.231 & 1.285 & 2.454 & 1.248 & 1.284 & 70 & 90 & 90 \\ 
  0.95 & 0.70 & 2.395 & 1.305 & 1.209 & 2.397 & 1.319 & 1.212 & 60 & 100 & 70 \\ 
  0.95 & 0.90 & 2.552 & 1.263 & 1.329 & 2.550 & 1.262 & 1.330 & 60 & 40 & 30 \\ 
  0.99 & 0.10 & 5.764 & 5.785 & 6.276 & 8.517 & 5.841 & 6.332 & 100 & 30 & 30 \\ 
  0.99 & 0.30 & 6.421 & 5.959 & 6.253 & 8.115 & 6.243 & 6.439 & 90 & 80 & 60 \\ 
  0.99 & 0.50 & 8.043 & 6.173 & 6.515 & 8.881 & 6.431 & 6.927 & 100 & 70 & 70 \\ 
  0.99 & 0.70 & 7.100 & 6.416 & 6.398 & 8.502 & 6.423 & 6.485 & 70 & 60 & 80 \\ 
  0.99 & 0.90 & 8.793 & 6.260 & 6.567 & 9.379 & 6.261 & 6.575 & 50 & 50 & 70 \\ 
  0.999 & 0.10 & 72.737 & 59.497 & 59.173 & 81.465 & 59.410 & 58.982 & 30 & 70 & 90 \\ 
  0.999 & 0.30 & 81.618 & 54.574 & 57.523 & 81.602 & 54.665 & 57.695 & 20 & 90 & 70 \\ 
  0.999 & 0.50 & 66.193 & 56.857 & 60.915 & 75.884 & 56.918 & 61.286 & 40 & 80 & 60 \\ 
  0.999 & 0.70 & 78.040 & 56.413 & 54.720 & 82.003 & 56.489 & 56.527 & 20 & 60 & 70 \\ 
  0.999 & 0.90 & 82.441 & 58.920 & 56.798 & 82.436 & 59.044 & 56.925 & 10 & 80 & 80 \\ 
  \hline
\end{tabular}
\caption{Comparing standardized squared Frobenius norm difference between true and estimated components. Defaults: n=200 for training set, $n=200$ for test set, $k=2$, $p=200$ for each dataset, joint and individual components have equal weight, and all ranks are 1.}
\label{tab:JIVE_norm}
\end{table}

\begin{table}
\centering
\footnotesize
\begin{tabular}{ccc|cccc}
  \hline
  Rank J  & Rank A1 & Rank A2 & \% sJIVE J & \% JIVE J & \% sJIVE A & \% JIVE A  \\ 
 \hline
1 & 1 & 1 & 100 & 90 & 20 & 100 \\ 
  1 & 1 & 2 & 70 & 70 & 20 & 70 \\ 
  1 & 1 & 3 & 50 & 40 & 0 & 100 \\ 
  1 & 1 & 4 & 70 & 80 & 0 & 40 \\ 
  1 & 2 & 2 & 50 & 90 & 10 & 70 \\ 
  1 & 2 & 3 & 20 & 90 & 0 & 80 \\ 
  1 & 2 & 4 & 60 & 50 & 0 & 30 \\ 
  1 & 3 & 3 & 20 & 70 & 0 & 60 \\ 
  1 & 3 & 4 & 80 & 70 & 0 & 40 \\ 
  1 & 4 & 4 & 30 & 50 & 10 & 10 \\ 
  2 & 1 & 1 & 50 & 0 & 60 & 70 \\ 
  2 & 1 & 2 & 20 & 0 & 10 & 40 \\ 
  2 & 1 & 3 & 40 & 0 & 0 & 40 \\ 
  2 & 1 & 4 & 0 & 10 & 10 & 0 \\ 
  2 & 2 & 2 & 10 & 0 & 10 & 70 \\ 
  2 & 2 & 3 & 50 & 0 & 10 & 40 \\ 
  2 & 2 & 4 & 0 & 10 & 0 & 0 \\ 
  2 & 3 & 3 & 10 & 0 & 0 & 20 \\ 
  2 & 3 & 4 & 20 & 0 & 0 & 10 \\ 
  2 & 4 & 4 & 30 & 0 & 0 & 10 \\ 
  3 & 1 & 1 & 0 & 0 & 20 & 0 \\ 
  3 & 1 & 2 & 60 & 0 & 10 & 0 \\ 
  3 & 1 & 3 & 30 & 0 & 10 & 0 \\ 
  3 & 1 & 4 & 20 & 0 & 0 & 0 \\ 
  3 & 2 & 2 & 20 & 0 & 10 & 0 \\ 
  3 & 2 & 3 & 30 & 10 & 10 & 0 \\ 
  3 & 2 & 4 & 20 & 10 & 0 & 0 \\ 
  3 & 3 & 3 & 30 & 0 & 10 & 10 \\ 
  3 & 3 & 4 & 20 & 0 & 0 & 30 \\ 
  3 & 4 & 4 & 10 & 0 & 0 & 0 \\ 
  4 & 1 & 1 & 0 & 0 & 30 & 0 \\ 
  4 & 1 & 2 & 20 & 0 & 10 & 0 \\ 
  4 & 1 & 3 & 0 & 0 & 10 & 0 \\ 
  4 & 1 & 4 & 0 & 0 & 0 & 0 \\ 
  4 & 2 & 2 & 0 & 0 & 0 & 0 \\ 
  4 & 2 & 3 & 30 & 10 & 0 & 0 \\ 
  4 & 2 & 4 & 0 & 0 & 10 & 0 \\ 
  4 & 3 & 3 & 10 & 0 & 0 & 0 \\ 
  4 & 3 & 4 & 10 & 0 & 0 & 0 \\ 
  4 & 4 & 4 & 0 & 0 & 0 & 0 \\ 
   \hline
\end{tabular}
\caption{Comparing rank selection techniques of sJIVE and JIVE. Each row is the average of 10 simulations. Defaults: $n=100$ $k=2$, $p=100$ for each dataset, joint and individual components have equal weight, 50\% of variation in $X$ is noise, 10\% of variation in $Y$ is noise, and all ranks are 1.}
\label{tab:JIVE_rank}
\end{table}

\begin{table}
\centering
\footnotesize
\begin{tabular}{l|ccc}
  \hline
  & Times Correctly & Times & Times \\
  & Specified & Underestimated & Overestimated \\
 \hline
 sJIVE CV  &&&\\
rank Selection  &&&\\
- J rank & 109 (27.3\%) & 208 (52.0\%) & 83 (20.8\%) \\
- A1 rank & 113 (28.3\%) & 180 (45.0\%) & 107 (26.8\%) \\
- A2 rank & 105 (26.3\%) & 197 (49.3\%) & 98 (24.5\%) \\
  JIVE Permutation  &&&\\
 rank Selection &&&\\
- J rank & 75 (18.8\%) & 322 (80.5\%) & 3 (0.8\%) \\
- A1 rank & 156 (39.0\%) & 10 (2.5\%) & 234 (58.5\%) \\
- A2 rank & 170 (42.5\%) & 29 (7.2\%) & 201 (50.2\%) \\
   \hline
\end{tabular}
\caption{Comparing rank selection techniques of sJIVE and JIVE. 400 total simulations were run with true ranks ranging from 1 to 4.}
\label{tab:JIVE_rank2}
\end{table}

\begin{table}
\centering
\footnotesize
\begin{tabular}{lccccccc}
  \hline
 & sJIVE & JIVE & Concatenated & Individual & Individual & CVR & \% sJIVE \\ 
Scenario & & Predict & PCA & PCA 1 & PCA 2 & & Wins \\
  \hline
Default & 0.1272 & 0.1323 & 0.1382 & 0.4242 & 0.4432 & 0.2151 & 90 \\ 
  High Dimensional (p=500) & 0.1047 & 0.1171 & 0.1238 & 0.4738 & 0.3576 & 0.1681 & 90 \\ 
  Large X error (0.99) & 0.4763 & 0.5532 & 0.6115 & 0.7173 & 0.7476 & 0.7590 & 100 \\ 
  K=4 & 0.2951 & 0.3068 & 0.3257 & 0.7115 & 0.6925 & -- & 80 \\ 
  Large Joint weight & 0.0437 & 0.0429 & 0.0426 & 0.0662 & 0.0656 & 0.0987 & 20 \\ 
  Large Individual weight & 0.0923 & 0.0926 & 0.0985 & 0.5782 & 0.5412 & 0.2017 & 60 \\ 
  All ranks=10 & 0.6761 & 0.7174 & 0.7612 & 0.8261 & 0.8492 & 0.9969 & 100 \\ 
  Ranks=10, but only first & 0.7419 & 0.7891 & 0.8040 & 0.8759 & 0.8909 & 1.0240 & 80 \\
  rank predicts Y &&&&&&& \\
   \hline
\end{tabular}
\caption{Comparing test MSE of sJIVE to other existing methods. Defaults: $n=200$ for training and test set, $k=2$, $p=200$ for each dataset, joint and individual components have equal weight, 90\% of variation in $X$ is noise, 1\% of variation in $Y$ is noise, and all ranks are 1.}
\label{tab:Compare}
\end{table}

\begin{table}
\centering
\footnotesize
\begin{tabular}{l|cc}
  \hline
   Model & Time & Test MSE \\
   \hline 
   sJIVE & 66.7 min & 0.6980 \\
   JIVE-predict & 34.6 min & 0.6991 \\
   Concatenated PCA & 4.5 sec & 0.7477 \\
   Individual PCA 1 & 4.2 sec & 0.7832 \\
   Individual PCA 2 & 0.4 sec & 0.7610 \\
   CVR & 110.9 hrs & 0.9805 \\
   \hline
\end{tabular}
 \caption{Comparing test MSE of sJIVE to other existing methods on the COPDGene data.}
\label{tab:COPD}
\end{table}

\begin{table}
\centering
\footnotesize
\begin{tabular}{l|ccc}
  \hline
   Predictor & Rank & Partial $R^2$ & P-value \\
   \hline 
   Joint Component & 1 & 0.069 & 0.001 \\
   RNAseq & 27 & 0.191 & 0.217 \\
   Proteomics & 24 & 0.335 & $<$0.001 \\
   \hline
\end{tabular}
 \caption{Results to sJIVE model on COPDGene data.}
\label{tab:COPD_sJIVE}
\end{table}

\begin{table}
\centering
\footnotesize
\begin{tabular}{cccccc}
  \hline
   & Pathway & Number of Genes & Ratio & P-value & FDR \\
   \hline 
   1 & Glucagon signaling pathway & 18 & 3.58 & $4.70 \times 10^{-7}$ & $1.05 \times 10^{-4}$ \\
   2 & Dopaminergic synapse & 20 & 3.23 & $3.23 \times 10^{-6}$ & $4.36 \times 10^{-4}$ \\
   3 & Cholinergic synapse & 20 & 3.00 & $3.06 \times 10^{-5}$ & $2.29 \times 10^{-3}$ \\
   4 & Wnt signaling pathway & 32 & 2.45 & $7.15 \times 10^{-5}$ & $3.27 \times 10^{-3}$ \\
   5 & B cell receptor signaling pathway & 24 & 2.69 & $8.43 \times 10^{-5}$ & $3.27 \times 10^{-3}$ \\
   6 & Chemokine signaling pathway & 74 & 1.87 & $1.02 \times 10^{-4}$ & $3.27 \times 10^{-3}$ \\
   7 & AGE-RAGE signaling pathways in diabetic complications & 51 & 2.08 & $1.05 \times 10^{-4}$ & $3.27 \times 10^{-3}$ \\
   8 & VEGF signaling pathway & 33 & 2.37 & $1.19 \times 10^{-4}$ & $3.27 \times 10^{-3}$ \\
   9 & Circadian entrainment & 10 & 3.69 & $1.31 \times 10^{-4}$ & $3.27 \times 10^{-3}$ \\
   10 & Insulin signaling pathway & 31 & 2.38 & $1.86 \times 10^{-4}$ & $4.16 \times 10^{-3}$ \\
   \hline
\end{tabular}
 \caption{Results to pathway analysis of Proteomic dataset.}
\label{tab:COPD_pathway}
\end{table}

\clearpage
\appendix
\section*{Appendix}
This appendix provides further details and validation of the sJIVE method. In Section \ref{section:7.1}, we confirm the uniqueness of the solution under orthogonality. In Section \ref{section:7.2}, we give additional details on the cross-validation approach for rank selection.  In Section \ref{section:7.3}, we give additional details on how the simulation data was generated. In Section \ref{section:6.4}, we describe how we reduce dimensionality to increase computational efficiency.

\section{Proof of Theorem 1}
\label{section:7.1}
Here, we provide a proof for Theorem 1 of the main article.  It follows from Lemma 1 of Feng et al. (2018) \citep{feng2018angle}, which is analogous to Theorem 1.1 of Lock et al. (2013) \citep{JIVE}, that a decomposition satisfying conditions 1., 2., and 3. of Theorem 1 exists and is unique for $\{\X_1,\hdots,\X_k\}$.  Further, because $\tilde{\y} \in \row(\tilde{\X}) \subset (\row(J)+\row(A))$ it follows that $\y = \j_y+\a_y$ where $\j_y \in \row(J)$ and $\a_y \in \row(\A)$ (condition 4.), and because $\row(J) \perp \row(A)$ $\j_y$ and $\a_y$ are uniquely defined.   

\section{Cross-Validation Rank Selection Algorithm}
\label{section:7.2}

sJIVE selects ranks using a forward selection 5-fold CV algorithm. In the pseudocode below, each time the algorithm instructs to run 5-fold CV, the data was split into 5 folds and sJIVE was fit to 4 of them. The fitted model was then used to predict the outcome for the left-out fold, and the MSE is recorded by comparing the estimated and true $\Y$ values. This is repeated for each of the 5 folds.

\textbf{\hspace{-.5cm} Pseudocode for sJIVE rank selection} \\
\framebox{

\parbox[t][8.5cm]{13.5cm}{

\addvspace{0.1cm} \flushleft
\begin{enumerate}
     \item Initialize $r_J$ and $r_i$ $\forall i=1,\dots,k$ to 0
     \item Run 5-fold CV. Record the average test MSE of each fold and label it $MSE_{best}$
     \item Let $r_J = r_J + 1$ and run 5-fold CV, recording the average test MSE of each fold
     \item Return $r_J$ back to $r_J-1$ 
     \item For $i=1,\dots,K$
     \begin{enumerate}
         \item Let $r_i = r_i+1$ and run 5-fold CV, recording the average test MSE of each fold
         \item Return $r_i$ back to $r_i-1$
     \end{enumerate}
     \item Determine which rank increase led to the largest reduction of test MSE. Permanently increase that rank by 1, and set $MSE_{best}$ to its new lower value.
     \item Repeat steps 3-6 until all rank increases lead to higher $MSE_{best}$ values
\end{enumerate}
} 
}

\section{Generating Simulated Datasets}
\label{section:7.3}

In this section, we will describe how the datasets were simulated. Our function for generating data allowed us to input the following: the number of datasets, $k$; the number of predictors in each dataset, $p=(p_1,\cdots,p_k)$; the number of observations, $n$; the joint and individual ranks, $r_{J}$ and $r_{A}=(r_{A_1}, \cdots, r_{A_k})$; the weight of the joint and individual signals, $w_J$ and $w_A$; the proportion of variance in $\mathbf{X}_i$, $i=1,\cdots,k$ that contributes to error, $X_{err}$; the proportion of $\mathbf{y}$ variance contributed to error, $Y_{err}$; and the proportion of the ranks that are predictive of $\mathbf{y}$, $r_{prop}$.

Define the following: 
\begin{itemize}
    \item $\mathbf{U}_i =
    \begin{bmatrix}
    runif(0.5,1) 
    \end{bmatrix}_{\lbrace p_i \times r_J \rbrace}$
    \item $\boldsymbol{\theta}_1 =  
    \begin{bmatrix}
    runif(0.5,1) & 0
    \end{bmatrix}_{\lbrace 1 \times r_J \rbrace}$ with the first $r_{prop} \times r_J$ values being non-zero
    \item take QR decomposition of $\begin{bmatrix}
    \mathbf{U}_1 \\ \vdots \\ \mathbf{U}_k \\ \boldsymbol{\theta}_1
    \end{bmatrix}_{\lbrace \sum p_i +1 \times r_J \rbrace}$ for new $\mathbf{U}_i$ and $\boldsymbol{\theta}_1$ values
    \item $\mathbf{S}_J =$ diag($w_J)_{\lbrace r_J \times r_J \rbrace}
    \begin{bmatrix}
    rnorm(0,1)
    \end{bmatrix}_{\lbrace r_J \times n \rbrace}$
\end{itemize}
for each dataset $i=1,...,K$
\begin{itemize}
     \item $\mathbf{W}_i =
    \begin{bmatrix}
    runif(0.5,1)
    \end{bmatrix}_{\lbrace p_i \times r_{A_i} \rbrace}$
    \item $\boldsymbol{\theta}_{2i} =  
    \begin{bmatrix}
    runif(0.5,1) & 0
    \end{bmatrix}_{\lbrace 1 \times r_{A_i} \rbrace}$ with the first $r_{prop} \times r_{A_i}$ values being non-zero
    \item take QR decomposition of $\begin{bmatrix}
    \mathbf{W}_i \\ \boldsymbol{\theta}_{2i}
    \end{bmatrix}_{\lbrace p_i +1 \times r_{A_i} \rbrace}$ for new $\mathbf{W}_i$ and $\boldsymbol{\theta}_{2i}$ values
    \item $\mathbf{S}_i =$ diag($w_A)_{\lbrace r_{A_i} \times r_{A_i} \rbrace}
    \begin{bmatrix}
    rnorm(0,1)
    \end{bmatrix}_{\lbrace r_{A_i} \times n \rbrace} \cdot \Big( I_{\lbrace n \times n \rbrace} - P_{\mathbf{S}_J} \Big)$
    \begin{itemize}
        \item where $P_{\mathbf{S}_J}= \mathbf{S}_J^T (\mathbf{S}_J \mathbf{S}_J^T)^{-1} \mathbf{S}_J$ to force orthogonality between $\mathbf{S}_J$ and $\mathbf{S}_i$
    \end{itemize}
\end{itemize}
We then can calculate $\mathbf{X}$ and $\mathbf{y}$
\begin{itemize}
    \item $\mathbf{X}_i = \mathbf{U}_i \mathbf{S}_J + \mathbf{W}_i \mathbf{S}_i + \E_i$ where $\E_i$ is normal with variance s.t. $var(\E_i)/var(\X_i)=X_{err}$
    \item $\Y = \boldsymbol{\theta}_1 \mathbf{S}_J + \sum_i \boldsymbol{\theta}_{2i} \mathbf{S}_i + \E_Y$ where $\E_Y$ is normal with variance s.t. $var(\E_Y)/var(\Y)=Y_{err}$

\item Scale $\mathbf{X}$ and $\mathbf{y}$ to have variance=1

\item Scale components to force $\Vert[ \mathbf{U}^T \boldsymbol{\theta}_1^T ]^T \Vert^2_F = \Vert [ \mathbf{W}^T_i \boldsymbol{\theta}_{2i}^T ]^T \Vert^2_F = 1 $ $\forall i=1,\dots,k$ for uniqueness
\end{itemize}
Return $\mathbf{X}_i$, $i=1,\dots,k$ and $\mathbf{y}$.

\section{Reducing dimensionality}
\label{section:6.4}

Though computation time relies on a variety of factors, data dimensions and rank selection are the main drivers. The computation times displayed in Table 6 of the main article were conducted on a 2.4 GHz computer with 8 GB RAM. In high-dimensional scenarios, JIVE maps the data into a lower dimension space before running its optimization function in increase efficiency. 

sJIVE uses this same technique. Consider a high-dimensional scenario when $p_i>>n$ for the $i$th dataset. Then let $\textbf{X}_i$ be the $p_i \times n$ data matrix. Prior to running the optimization function, map $\textbf{X}_i$ to an $n \times n$ space using SVD, i.e.,

\begin{align*}
    SVD(\textbf{X}_i) &= UDV^T \\
    \textbf{X}_i^\perp &= DV^T
\end{align*}

This transformation preserves covariance and Euclidean distance between columns in $\textbf{X}_i$. By implementing this reduction in dimensionality, computation time can significantly decrease. For example in our COPDGene application, sJIVE took 52.6 hours to run without data reduction, but only 66 minutes after utilizing this transformation.

After optimizing the function and obtaining estimates for each of the joint and individual components, $\textbf{X}_i^\perp$ can be mapped back to the original space by multiplying the left singular vectors, $U$, by the estimated loadings, $\textbf{U}_i^\perp$ and $\textbf{W}_i^\perp$ $i=1,\dots,k$. The scores, $\textbf{S}_J$ and $\textbf{S}_i$, and the $\boldsymbol{\theta}$ coefficients do not require any transformation.

\end{document}